\documentclass{article}

\usepackage{microtype}
\usepackage{graphicx}
\usepackage{subfigure}
\usepackage{booktabs} %
\usepackage{hyperref}

\usepackage[accepted]{icml}

\usepackage{amsmath}
\usepackage{amssymb}
\usepackage{mathtools}
\usepackage{amsthm}

\usepackage[capitalize,noabbrev]{cleveref}

\theoremstyle{plain}

\theoremstyle{definition}

\theoremstyle{remark}

\usepackage[textsize=tiny]{todonotes}

\usepackage[inline]{enumitem}

\usepackage{float}
\icmltitlerunning{Early Stopping Tabular In-Context Learning}

\begin{document}

\twocolumn[
\icmltitle{Early Stopping Tabular In-Context Learning}

\icmlsetsymbol{equal}{*}

\begin{icmlauthorlist}
\icmlauthor{Jaris Küken}{ufr,eliza}
\icmlauthor{Lennart Purucker}{ufr}
\icmlauthor{Frank Hutter}{prior,ellis,ufr}
\end{icmlauthorlist}

\icmlaffiliation{ufr}{Department of Computer Science, University of Freiburg, Freiburg, Germany}
\icmlaffiliation{eliza}{Zuse School ELIZA, Darmstadt, Germany}
\icmlaffiliation{ellis}{ELLIS Institute, Tübingen, Germany}
\icmlaffiliation{prior}{Prior Labs, Freiburg, Germany}

\icmlcorrespondingauthor{Jaris Küken}{kuekenj@cs.uni-freiburg.de}

\icmlkeywords{Machine Learning, ICML}

\vskip 0.3in
]

\printAffiliationsAndNotice{\icmlEqualContribution} %

\begin{abstract}
Tabular foundation models have shown strong performance across various tabular learning tasks via in-context learning, offering robust generalization without any downstream finetuning. However, their inference-time costs remain high, particularly for larger datasets. 
To address this, we propose early-stopping the in-context learning process.
We achieve this by dynamically evaluating whether to stop in-context learning after each Transformer encoder layer. 
Once stopped, we decode the embedding using a pre-trained layer-wise decoder. 
Experiments across 34 small classification tasks size show that early stopping in-context learning accelerates inference by up to $\times1.3$ with negligible degradation in predictive performance. To assess scalability, we further evaluate our method on five larger classification tasks, achieving speedups of up to $\times2.2$. Our results demonstrate the potential of early exiting as an effective and practical strategy for improving the efficiency of tabular in-context learning.
\end{abstract}

\section{Introduction}\label{sec:introduction}
Tabular data is widely present across domains such as finance or healthcare \citep{borisov-nnls22a, vanbreugel-icml24a}, many of which involve time-critical decision making. Traditionally, tree-based models like XGBoost~\citep{chen-kdd16a}, LightGBM~\citep{ke-neurips17a} or CatBoost~\citep{prokhorenkova-neurips18a} have demonstrated strong performance on tabular problems, often surpassing deep learning approaches \citep{grinsztajn-neurips22a}. However, these models typically require extensive training and hyperparameter tuning for each downstream task. \\
Recent research has shifted towards tabular foundation models (TFMs) \citep{vanbreugel-icml24a}, many of which leverage in-context learning \citep{brown-neurips20a} to enable fast adaption with minimal or no task-specific training. Notable examples include TabPFN~\citep{hollmann-nature25a, hollmann-iclr23a}, TabICL~\citep{qu_tabicl_2025} or TabDPT~\citep{ma_tabdpt_2024}. These models build on top of the Transformer architecture \citep{vaswani-neurips17a} to condition the context data and perform inference directly on new tasks. However, the self-attention mechanism scales quadratically with context size, making inference costly in TFMs. \\
We investigate whether the in-context learning in TFMs can be stopped early—without significant loss in performance—through an entropy-based early exiting strategy. Specifically, we pre-train lightweight decoders at each Transformer layer on synthetic data and use them to probe the test sample during inference. Based on the entropy of the prediction, we determine whether to exit early. This enables inference speedups of up to $\times1.3$ on small datasets and up to $\times2.2$ on larger tasks, with minimal degradation in performance and without the need for any downstream task-specific finetuning. Thus, we preserve a key advantage of TFMs:
strong predictive performance without downstream task-specific finetuning.
While early-exit strategies have been explored in natural language processing \citep{xin_deebert_2020, liu_fastbert_2020, hou_dynabert_2020} and vision \citep{teerapittayanon_branchynet_2016}, they remain largely underexplored in the context of in-context learning. Our work aims to close this gap and highlights the potential of early exiting as a practical tool for improving TFM efficiency.
\textbf{Our contributions are:}
\begin{enumerate*}[label=\textbf{(\Roman*)}]
    \item We introduce a simple yet effective entropy-based early-stopping mechanism for tabular in-context learning.
    \item We demonstrate that inference of TFMs can be sped up by up to $\times1.3$ on small tasks and $\times2.2$ on larger tasks with negligible loss in predictive performance.
    \item Unlike prior early-exit methods, our approach does not require task-specific finetuning, maintaining the advantages of in-context learning of TFMs. 
\end{enumerate*}

\section{Related Work}\label{sec:relatedwork}
\textbf{Improving Transformer Efficiency.}\quad
Improving the computational efficiency of Transformer models \citep{vaswani-neurips17a} has been widely studied, particularly natural language processing. Early exit strategies have been introduced for encoder-based architectures like BERT~\citep{devlin-acl19a}, allowing inference to terminate at intermediate layers without traversing the full model. Notable examples include DynaBERT~\citep{hou_dynabert_2020}, FastBERT~\citep{liu_fastbert_2020} or CascadeBERT~\citep{li_cascadebert_2021}. Additional efficiency-oriented approaches such as quantization \citep{kim_i-bert_2021, shen_q-bert_2020, zafrir_q8bert_2019} and knowledge distillation \citep{hinton_distilling_2015, wang_minilm_2020, sanh_distilbert_2020}, have also been proposed, though they typically involve model compression rather then dynamic inference control and are therefore related to our work.
Our method is most related to early-exit architectures like DeeBERT~\citep{xin_deebert_2020} and BranchyNet~\citep{teerapittayanon_branchynet_2016}, but differs in its application to in-context learning. In contrast to prior work, which often requires task-specific finetuning to calibrate exit decisions, our approach leverages a pretrained, fixed exit mechanism that enables efficient inference.

\textbf{Tabular In-Context Learning.}\quad
In-context learning (ICL) \citep{brown-neurips20a} has recently been adapted to  tabular data through several distinct approaches. One prominent line of work interprets ICL as approximate Bayesian inference over tabular tasks \citep{xie_explanation_2021, muller-iclr22a, reuter_can_2025}, leading to the development of Transformer-based foundation models such as TabPFN~\citep{hollmann-iclr23a, hollmann-nature25a}, TabICL~\citep{qu_tabicl_2025} or TabDPT~\citep{ma_tabdpt_2024}. 

\section{Method}\label{sec:method}
\begin{figure*}[ht]
    \centering
    \includegraphics[width=\textwidth]{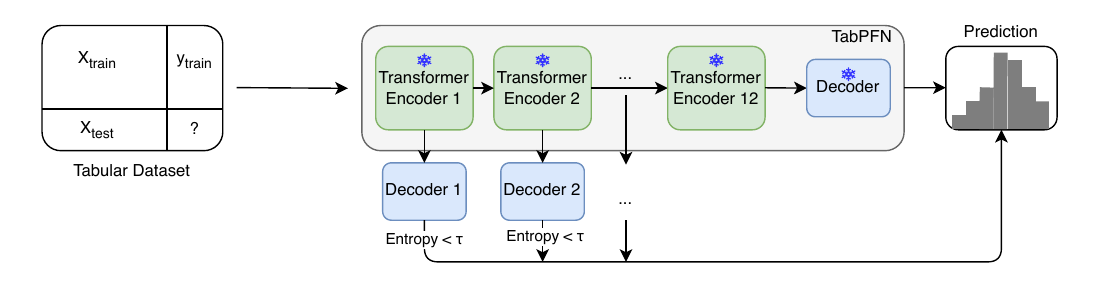}
    \caption{\textbf{Early Exit Strategy for TabPFN.} We extend TabPFN with an early-exit mechanism for in-context learning. For each Transformer layer, we pre-train a dedicated decoder on synthetically generated data from the prior (blue) while keeping the Transformer layers (green) and the final decoder frozen. During inference, all test samples are passed through each decoder in sequence. If the average prediction entropy falls below a predefined threshold $\tau$, the forward pass is terminated early and predictions are returned.}
    \label{fig:setup}
\end{figure*}

We propose a straightforward yet effective early-exit mechanism for tabular in-context learning.
Our method can be used with any TFM; for this study, we use TabPFNv2~\citep{hollmann-nature25a}. Our approach introduces pre-trained, layer-specific decoders into the Transformer~\citep{vaswani-neurips17a} architecture, enabling dynamic termination of the forward pass during inference. At each layer, we decide whether to exist early based on prediction entropy, as illustrated in Figure~\ref{fig:setup}. 
Crucially, our method requires no task-specific finetuning and preserves the zero-shot capabilities of the base model. All exit points are trained on synthetically generated data from a prior.

\textbf{Method Motivation.}\quad
We investigate whether intermediate layers of TabPFN can produce useful predictions prior to the final layer. As shown in Figure~\ref{fig:final_v_indiv_pmlb}, activations from earlier layers, when passed through the final decoder, already yield non-trivial performance—albeit typically suboptimal compared to the last layer. This indicates that significant predictive information emerges well before the end of the forward pass. To better leverage these early representations, we pre-train dedicated decoders for each Transformer layer, enabling flexible early exits.

\textbf{Layer-Specific Decoder Pretraining.}\quad
We modify the original TabPFN~\citep{hollmann-nature25a} architecture by attaching a decoder to each encoder layer, as illustrated in Figure~\ref{fig:setup}. Each intermediate decoder shares the architecture of the original TabPFN decoder, but is trained independently.
To enable generalization across tasks, we pre-train each decoder on synthetic datasets sampled from the prior introduced by \citet{muller-arxiv23a}\footnote{We adapt the pre-training setup specified in: \url{https://github.com/microsoft/ticl}}.
During training, the Transformer backbone is frozen, and only the decoder corresponding to the current exit layer is optimized. For the final layer, we retain the original TabPFN decoder. In total, each decoder is trained on approximately 820,000 synthetic datasets. Details of the data generation and training procedure are provided in Appendix~\ref{app:decoder_training}.

\textbf{Early Stopping Based on Prediction Entropy.}\quad
At inference time, we perform early stopping based on the predictive uncertainty of each layer's decoder, similar to BranchyNet~\citep{teerapittayanon_branchynet_2016} and DeeBERT~\citep{xin_deebert_2020}. After each Transformer layer, all test samples are passed through the corresponding decoder, and we compute the entropy of their predictive distributions. We then average the entropies across the test set and compare the result to a predefined threshold $\tau$.
If the average entropy falls below $\tau$, the model exits early and outputs the corresponding predictions. Otherwise, the forward pass continues to the next layer. The threshold $\tau$ is a dataset specific, tunable hyperparameter. Importantly, as all decoders are pre-trained, no adaption to downstream tasks is needed, thus making our method a cheap and efficient addition to the standard forward pass of TabPFN. A detailed overview of our proposed algorithm is specified Appendix~\ref{app:alg_details}.

\section{Experiments}\label{sec:experiments}
We evaluate our method on two categories of classification tasks to assess its effectiveness across dataset sizes. First, we benchmark on 34 small-scale binary classification tasks from the PMLB-mini suite~\citep{knauer_pmlbmini_2024} containing datasets with up to 500 samples\footnote{All experiments are run on a single NVIDIA RTX 2080 GPU}. 
Second, we assess scalability and inference efficiency on five larger classification datasets containing up to 5,000 samples, where TabPFN's inference latency becomes more critical due to the quadratic complexity of the self-attention mechanism. We specify all datasets in Appendix~\ref{app:datasets}.
For all experiments, we report mean performance over 10-fold cross-validation. We report details on the hyperparameters for the TabPFN evaluation in Appendix~\ref{app:hyperparameters}.

\textbf{Performance of Intermediate Layers.}\quad
We investigate the predictive capacity of intermediate Transformer layers in TabPFN. To do this, we pre-train a decoder for each layer as described in Section~\ref{sec:method}. For each benchmark dataset, we terminate the forward pass at each layer and pass the resulting representation through its corresponding decoder. We then measure the classification accuracy for each exit point.

\textbf{Early Stopping based on Entropy.}\quad
We evaluate the proposed entropy-based early-exit method as shown in Figure~\ref{fig:setup}. We test a range of five entropy thresholds $\tau$ to control early exit behavior. For each setting, we report the predictive accuracy and the average number of Transformer layers evaluated per inference run. This allows us to quantify the trade-off between accuracy and computational cost.

\textbf{Scalability to Larger Datasets.}\quad
To further evaluate the scalability of our approach, we test it on larger datasets containing up to 5.000 samples—settings where TabPFN's inference latency becomes particularly critical. Importantly, since our decoders are pre-trained only on synthetic datasets with up to $~\sim 1.000$ samples, this also serves as a test of the method’s ability to generalize to larger-scale tasks.

\section{Results}\label{sec:results}
\begin{figure}[h]
    \centering
    \includegraphics[width=\linewidth]{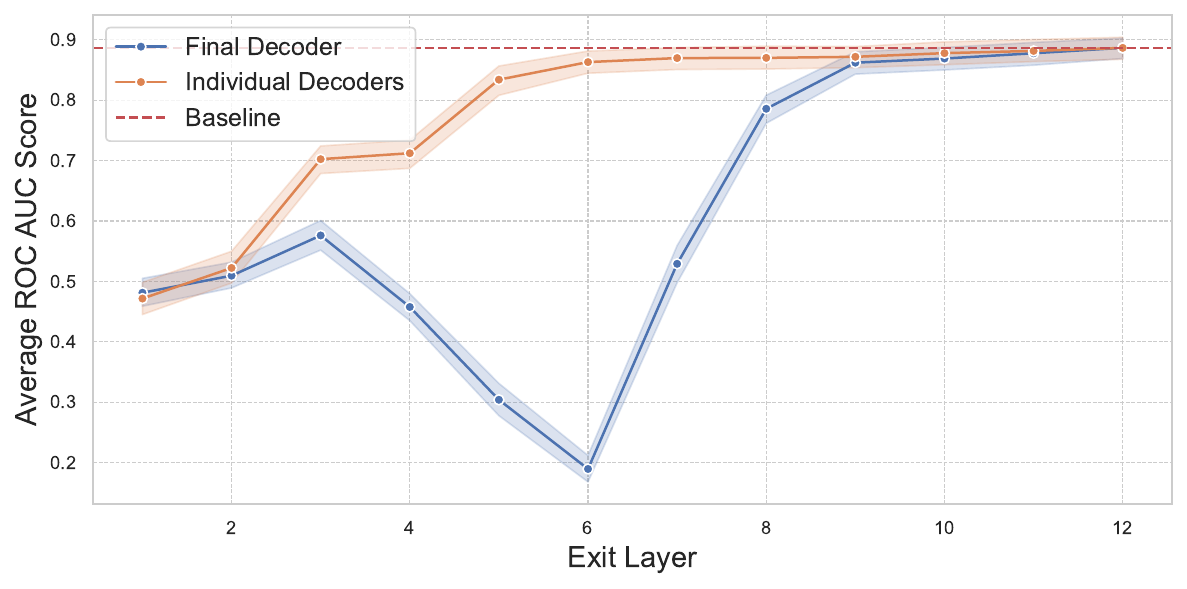}
    \caption{\textbf{Performance per Layer over 34 Small Datasets with Individual Decoders and Final Decoder.} We report the average ROC AUC score across 34 small-scale datasets, averaged over 10 folds, along with 95\% confidence intervals. We compare two exit strategies: using the original final decoder at intermediate layers (blue) and using individually pre-trained decoders specific to each layer (red). For reference, we also include the full TabPFN baseline, where inference proceeds through the entire Transformer (red dashed).
    Notably, intermediate layers already achieve strong performance, with layers as early as 5 matching the final layer’s accuracy when equipped with individual decoders. Moreover, these layer-wise decoders significantly outperform intermediate exits routed through the original final decoder.}
    \label{fig:final_v_indiv_pmlb}
\end{figure}

\begin{table*}
    \centering
    \caption{\textbf{Impact of Different Entropy Thresholds on Inference Performance.} We report ROC AUC, difference in wall-clock runtime (in seconds) compared to full TabPFN forward pass, and average exit layer across 34 small and 4 large classification datasets for 5 different entropy thresholds. All values are averaged across datasets.}
    \label{tab:threshold-performance}
    \resizebox{\textwidth}{!}{
    \begin{tabular}{lcccccc}
        \toprule
        & \multicolumn{6}{c}{\textbf{Small Datasets}} \\
        \cmidrule(lr){2-7}
        & Baseline & $\tau{=}0.1$ & $\tau{=}0.2$ & $\tau{=}0.3$ & $\tau{=}0.4$ & $\tau{=}0.5$ \\
        \midrule
            \textbf{ROC AUC} & $0.886 \pm 0.17$ &  $0.888 \pm 0.17 $ & $0.887 \pm 0.16$ & $0.883 \pm 0.17 $ & $0.873 \pm 0.18 $ & $0.861 \pm 0.19$ \\
            \textbf{Runtime $\Delta$ in (s)} & — & $0.004 \pm 0.00$ & $-0.013 \pm 0.00 $ & $-0.044 \pm 0.00$ & $-0.069 \pm 0.00$ & $-0.084 \pm 0.00$\\
            \textbf{Avg. Exit Layer} & 12.0 & 11.0 & 9.8 & 7.6 & 6.1 & 5.3 \\
        \bottomrule
    \end{tabular}}
    \resizebox{\textwidth}{!}{
    \begin{tabular}{lcccccc}
        & \multicolumn{6}{c}{\textbf{Large Datasets}} \\
        \cmidrule(lr){2-7}
        & Baseline & $\tau{=}0.1$ & $\tau{=}0.2$ & $\tau{=}0.3$ & $\tau{=}0.4$ & $\tau{=}0.5$ \\
        \midrule
            \textbf{ROC AUC} & $0.960 \pm 0.04 $ &  $0.957 \pm 0.04$ & $0.941 \pm 0.04$ & $0.923 \pm 0.04$ & $0.920\pm 0.04$ & $0.920 \pm 0.04$ \\
            \textbf{Runtime $\Delta$ in (s)} & — & $-0.550 \pm 0.51$ & $-1.194 \pm 0.36$ & $-1.369 \pm 0.45 $ & $-1.415 \pm 0.32$ & $-1.464 \pm 0.45$ \\
            \textbf{Avg. Exit Layer} & 12.0  & 9.48 & 7.02 & 5.42 & 5.0 & 5.0 \\
        \bottomrule
    \end{tabular}}
\end{table*}

\textbf{Intermediate Layers Enable Confident Early Exits.}\quad
Figure~\ref{fig:final_v_indiv_pmlb} shows the classification accuracy obtained when exiting TabPFN at different Transformer layers using their corresponding pre-trained decoders, compared to using only the final decoder. Results are averaged over 34 small-scale datasets from the PMLB-mini \citep{knauer_pmlbmini_2024} benchmark.
We observe that by pre-training individual decoders for each layer stabilizes performance across layers, with several intermediate layers reaching accuracy close to that of the final layer. This suggests that many inputs can be confidently predicted before completing the full forward pass, enabling faster inference with negligible loss in accuracy.
\begin{figure}[h]
    \centering
    \includegraphics[width=1.0\linewidth]{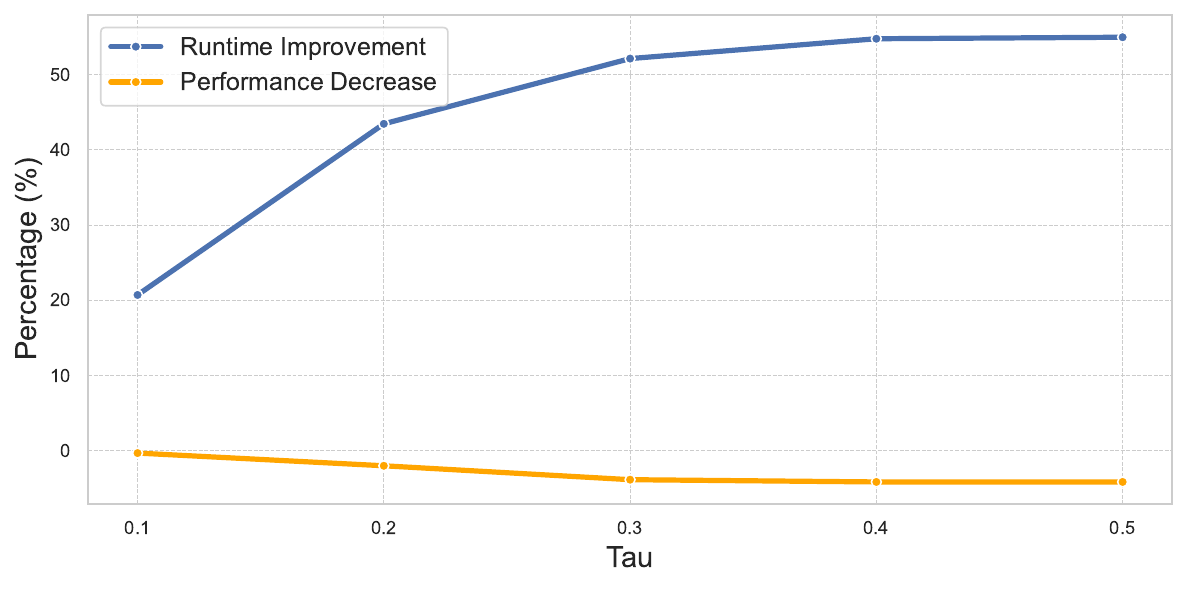}
    \caption{\textbf{Tradeoff between Improved Runtime and Decrease in Predictive Performance.} We report the relative improvements in runtime as well as the relative decrease in ROC AUC score averaged over 5 large datasets when early stopping based on entropy for 5 different entropy thresholds. It is trivial to see that a higher threshold $\tau$ leads to higher improvements in runtime, but also leads to a stronger decrease in terms of predictive performance. Notably, the gains in terms of relative runtime improvement strongly exceed the relative performance decrease.}
    \label{fig:nature-dynamic-stopping}
\end{figure}

\textbf{Entropy-Based Early Exit Trades Accuracy for Efficiency.}\quad
To validate our early exit strategy, we apply the entropy-based criterion across a range of thresholds $\tau$. Figure~\ref{fig:nature-dynamic-stopping} illustrates the trade-off between runtime savings and classification accuracy for large datasets, where inference cost is especially pressing.
On larger tasks, thresholds in the range $\tau \in [0.4, 0.5]$ yield up to $\times2.2$ faster inference with only small accuracy degradation. On small-scale datasets, thresholds between $\tau \in [0.3, 0.4]$ already achieve up to $\times1.3$ speedup with negligible loss in predictive performance as shown in Table~\ref{tab:threshold-performance}. Appendix~\ref{app:eval_details} contains further evaluation results on small datasets. Across tasks, this allows TabPFN to preserve its strong predictive accuracy while significantly improving efficiency.
Table~\ref{tab:threshold-performance} provides additional insights: the average exit layer, actual runtime savings in seconds, and corresponding accuracy deltas, across both small and large datasets for a fixed set of threshold values. These results confirm that our early-exit mechanism adapts flexibly to varying data scales, maintaining competitive predictive performance while offering tangible gains in inference efficiency. We report all discussed metrics, per-dataset in Appendix~\ref{app:detailed_results}.
These findings demonstrate that entropy-based early exiting offers a practical and cheap mechanism to accelerate inference in tabular in-context learning without sacrificing significant predictive quality.

\textbf{Generalization to Larger Datasets.}\quad
Table~\ref{tab:threshold-performance} reports detailed results on larger datasets. Early stopping demonstrates strong potential in this setting, yielding up to $\times2.2$ faster inference for thresholds in the range $\tau \in [0.4, 0.5]$. While performance degradation becomes more noticeable—up to $4\%$—compared to smaller datasets, it remains moderate relative to the efficiency gains. These results indicate that early stopping can substantially reduce inference costs even for large-scale tasks, and pave the way for more refined and adaptive strategies to maintain performance.
At the same time, these results highlight the ability of the individual decoders generalize effectively to larger datasets, despite being trained only on smaller-scale data. Appendix~\ref{app:eval_details} provides further insights into the evaluation of all benchmark datasets.

\section{Conclusion}\label{sec:conclusion}
We propose early-exit tabular in-context learning, unlocking dynamic inference stopping for the Tabular Foundation Model (TFM) TabPFN based on prediction entropy. By pre-training decoders for each Transformer layer on synthetic data—without the need for any task-specific finetuning—our method introduces minimal overhead while significantly improving inference efficiency. Experiments across both small and large classification benchmarks indicate that many predictions can be confidently made before reaching the final layer, achieving up to $\times1.3$ faster inference runtime with only marginal loss in accuracy on small datasets. On larger-scale tasks this effect is even more pronounced with inference being up to $\times2.2$ faster, while still maintaining near-optimal performance.
Our approach retains the zero-shot generalization and strong predictive performance of TFMs while mitigating the inference cost associated with larger input contexts. These results highlight the possibility of early exiting, progressing towards real-time inference for tabular in-context learning.

\clearpage

\section*{Acknowledgements}
L.P. acknowledges funding by the Deutsche Forschungsgemeinschaft (DFG, German Research Foundation) under SFB 1597 (SmallData), grant number 499552394; and funding by ELSA – European Lighthouse on Secure and Safe AI funded by the European Union under Grant Agreement No. 101070617.
Frank Hutter acknowledges the financial support of the Hector Foundation. Jaris Küken is supported by the Konrad Zuse School of Excellence in Learning and Intelligent Systems (ELIZA) through the DAAD program Konrad Zuse Schools of Excellence in Artificial Intelligence, sponsored by the Federal Ministry of Education and Research. Finally, we
thank the reviewers for their constructive feedback and contribution to improving the paper.
\bibliography{lib,proc,strings,custom}

\begin{thebibliography}{30}
\providecommand{\natexlab}[1]{#1}
\providecommand{\url}[1]{\texttt{#1}}
\expandafter\ifx\csname urlstyle\endcsname\relax
  \providecommand{\doi}[1]{doi: #1}\else
  \providecommand{\doi}{doi: \begingroup \urlstyle{rm}\Url}\fi

\bibitem[Borisov et~al.(2022)Borisov, Leemann, Se{\ss}ler, Haug, Pawelczyk, and Kasneci]{borisov-nnls22a}
Borisov, V., Leemann, T., Se{\ss}ler, K., Haug, J., Pawelczyk, M., and Kasneci, G.
\newblock Deep neural networks and tabular data: A survey.
\newblock \emph{IEEE Transactions on Neural Networks and Learning Systems}, 2022.

\bibitem[Brown et~al.(2020)Brown, Mann, Ryder, Subbiah, Kaplan, Dhariwal, Neelakantan, Shyam, Sastry, Askell, Agarwal, Herbert-Voss, Krueger, Henighan, Child, Ramesh, Ziegler, Wu, Winter, Hesse, Chen, Sigler, Litwin, Gray, Chess, Clark, Berner, McCandlish, Radford, Sutskever, and Amodei]{brown-neurips20a}
Brown, T., Mann, B., Ryder, N., Subbiah, M., Kaplan, J., Dhariwal, P., Neelakantan, A., Shyam, P., Sastry, G., Askell, A., Agarwal, S., Herbert-Voss, A., Krueger, G., Henighan, T., Child, R., Ramesh, A., Ziegler, D., Wu, J., Winter, C., Hesse, C., Chen, M., Sigler, E., Litwin, M., Gray, S., Chess, B., Clark, J., Berner, C., McCandlish, S., Radford, A., Sutskever, I., and Amodei, D.
\newblock Language models are few-shot learners.
\newblock In Larochelle, H., Ranzato, M., Hadsell, R., Balcan, M.-F., and Lin, H. (eds.), \emph{Proceedings of the 33rd International Conference on Advances in Neural Information Processing Systems ({N}eur{IPS}'20)}, pp.\  1877--1901, 2020.

\bibitem[Chen \& Guestrin(2016)Chen and Guestrin]{chen-kdd16a}
Chen, T. and Guestrin, C.
\newblock {XGBoost}: {A} scalable tree boosting system.
\newblock In Krishnapuram, B., Shah, M., Smola, A., Aggarwal, C., Shen, D., and Rastogi, R. (eds.), \emph{Proceedings of the 22nd {ACM} {SIGKDD} International Conference on Knowledge Discovery and Data Mining ({KDD}'16)}, pp.\  785--794, 2016.

\bibitem[Devlin et~al.(2019)Devlin, Chang, Lee, and Toutanova]{devlin-acl19a}
Devlin, J., Chang, M., Lee, K., and Toutanova, K.
\newblock {BERT}: Pre-training of deep bidirectional transformers for language understanding.
\newblock In Burstein, J., Doran, C., and Solorio, T. (eds.), \emph{Proceedings of the 2019 Conference of the North {A}merican Chapter of the Association for Computational Linguistics: Human Language Technologies}, pp.\  4171--4186. Association for Computational Linguistics, 2019.

\bibitem[Grinsztajn et~al.(2022)Grinsztajn, Oyallon, and Varoquaux]{grinsztajn-neurips22a}
Grinsztajn, L., Oyallon, E., and Varoquaux, G.
\newblock Why do tree-based models still outperform deep learning on typical tabular data?
\newblock In Koyejo, S., Mohamed, S., Agarwal, A., Belgrave, D., Cho, K., and Oh, A. (eds.), \emph{Proceedings of the 35th International Conference on Advances in Neural Information Processing Systems ({N}eur{IPS}'22)}, 2022.

\bibitem[Guyon et~al.(2017)Guyon, von Luxburg, Bengio, Wallach, Fergus, Vishwanathan, and Garnett]{nips17}
Guyon, I., von Luxburg, U., Bengio, S., Wallach, H., Fergus, R., Vishwanathan, S., and Garnett, R. (eds.).
\newblock \emph{Proceedings of the 31st International Conference on Advances in Neural Information Processing Systems ({N}eur{IPS}'17)}, 2017.

\bibitem[Hinton et~al.(2015)Hinton, Vinyals, and Dean]{hinton_distilling_2015}
Hinton, G.~E., Vinyals, O., and Dean, J.
\newblock Distilling the {Knowledge} in a {Neural} {Network}.
\newblock \emph{CoRR}, abs/1503.02531, 2015.
\newblock URL \url{http://arxiv.org/abs/1503.02531}.
\newblock arXiv: 1503.02531.

\bibitem[Hollmann et~al.(2023)Hollmann, M{\"u}ller, Eggensperger, and Hutter]{hollmann-iclr23a}
Hollmann, N., M{\"u}ller, S., Eggensperger, K., and Hutter, F.
\newblock Tab{PFN}: A transformer that solves small tabular classification problems in a second.
\newblock In \emph{The Eleventh International Conference on Learning Representations ({ICLR}'23)}. ICLR, 2023.
\newblock Published online: \url{iclr.cc}.

\bibitem[Hollmann et~al.(2025)Hollmann, M{\"u}ller, Purucker, Krishnakumar, K{\"o}rfer, Hoo, Schirrmeister, and Hutter]{hollmann-nature25a}
Hollmann, N., M{\"u}ller, S., Purucker, L., Krishnakumar, A., K{\"o}rfer, M., Hoo, S.~B., Schirrmeister, R.~T., and Hutter, F.
\newblock Accurate predictions on small data with a tabular foundation model.
\newblock \emph{Nature}, 637\penalty0 (8045):\penalty0 319--326, 2025.

\bibitem[Hou et~al.(2020)Hou, Huang, Shang, Jiang, Chen, and Liu]{hou_dynabert_2020}
Hou, L., Huang, Z., Shang, L., Jiang, X., Chen, X., and Liu, Q.
\newblock {DynaBERT}: {Dynamic} {BERT} with {Adaptive} {Width} and {Depth}.
\newblock In \emph{Advances in {Neural} {Information} {Processing} {Systems}}, volume~33, pp.\  9782--9793. Curran Associates, Inc., 2020.
\newblock URL \url{https://proceedings.neurips.cc/paper/2020/hash/6f5216f8d89b086c18298e043bfe48ed-Abstract.html}.

\bibitem[Ke et~al.(2017)Ke, Meng, Finley, Wang, Chen, Ma, Ye, and Liu]{ke-neurips17a}
Ke, G., Meng, Q., Finley, T., Wang, T., Chen, W., Ma, W., Ye, Q., and Liu, T.-Y.
\newblock Lightgbm: A highly efficient gradient boosting decision tree.
\newblock In  \citet{nips17}.

\bibitem[Kim et~al.(2021)Kim, Gholami, Yao, Mahoney, and Keutzer]{kim_i-bert_2021}
Kim, S., Gholami, A., Yao, Z., Mahoney, M.~W., and Keutzer, K.
\newblock I-{BERT}: {Integer}-only {BERT} {Quantization}.
\newblock \emph{International Conference on Machine Learning (Accepted)}, 2021.

\bibitem[Knauer et~al.(2024)Knauer, Grimm, and Rodner]{knauer_pmlbmini_2024}
Knauer, R., Grimm, M., and Rodner, E.
\newblock {PMLBmini}: {A} {Tabular} {Classification} {Benchmark} {Suite} for {Data}-{Scarce} {Applications}, September 2024.
\newblock URL \url{http://arxiv.org/abs/2409.01635}.
\newblock arXiv:2409.01635 [cs].

\bibitem[Li et~al.(2021)Li, Lin, Chen, Ren, Li, Zhou, and Sun]{li_cascadebert_2021}
Li, L., Lin, Y., Chen, D., Ren, S., Li, P., Zhou, J., and Sun, X.
\newblock {CascadeBERT}: {Accelerating} {Inference} of {Pre}-trained {Language} {Models} via {Calibrated} {Complete} {Models} {Cascade}.
\newblock In Moens, M.-F., Huang, X., Specia, L., and Yih, S. W.-t. (eds.), \emph{Findings of the {Association} for {Computational} {Linguistics}: {EMNLP} 2021}, pp.\  475--486, Punta Cana, Dominican Republic, November 2021. Association for Computational Linguistics.
\newblock \doi{10.18653/v1/2021.findings-emnlp.43}.
\newblock URL \url{https://aclanthology.org/2021.findings-emnlp.43/}.

\bibitem[Liu et~al.(2020)Liu, Zhou, Wang, Zhao, Deng, and Ju]{liu_fastbert_2020}
Liu, W., Zhou, P., Wang, Z., Zhao, Z., Deng, H., and Ju, Q.
\newblock {FastBERT}: a {Self}-distilling {BERT} with {Adaptive} {Inference} {Time}.
\newblock In Jurafsky, D., Chai, J., Schluter, N., and Tetreault, J. (eds.), \emph{Proceedings of the 58th {Annual} {Meeting} of the {Association} for {Computational} {Linguistics}}, pp.\  6035--6044, Online, July 2020. Association for Computational Linguistics.
\newblock \doi{10.18653/v1/2020.acl-main.537}.
\newblock URL \url{https://aclanthology.org/2020.acl-main.537/}.

\bibitem[Ma et~al.(2024)Ma, Thomas, Hosseinzadeh, Kamkari, Labach, Cresswell, Golestan, Yu, Volkovs, and Caterini]{ma_tabdpt_2024}
Ma, J., Thomas, V., Hosseinzadeh, R., Kamkari, H., Labach, A., Cresswell, J.~C., Golestan, K., Yu, G., Volkovs, M., and Caterini, A.~L.
\newblock {TabDPT}: {Scaling} {Tabular} {Foundation} {Models}, October 2024.
\newblock URL \url{http://arxiv.org/abs/2410.18164}.
\newblock arXiv:2410.18164 [cs].

\bibitem[M{\"u}ller et~al.(2023)M{\"u}ller, Curino, and Ramakrishnan]{muller-arxiv23a}
M{\"u}ller, A., Curino, C., and Ramakrishnan, R.
\newblock Mothernet: A foundational hypernetwork for tabular classification.
\newblock \emph{arXiv:2312.08598 [cs.LG]}, 2023.

\bibitem[M{\"u}ller et~al.(2022)M{\"u}ller, Hollmann, Arango, Grabocka, and Hutter]{muller-iclr22a}
M{\"u}ller, S., Hollmann, N., Arango, S., Grabocka, J., and Hutter, F.
\newblock Transformers can do {B}ayesian inference.
\newblock In \emph{The Tenth International Conference on Learning Representations ({ICLR}'22)}. ICLR, 2022.
\newblock Published online: \url{iclr.cc}.

\bibitem[Prokhorenkova et~al.(2018)Prokhorenkova, Gusev, Vorobev, Dorogush, and Gulin]{prokhorenkova-neurips18a}
Prokhorenkova, L., Gusev, G., Vorobev, A., Dorogush, A., and Gulin, A.
\newblock Catboost: Unbiased boosting with categorical features.
\newblock In Bengio, S., Wallach, H., Larochelle, H., Grauman, K., Cesa{-}Bianchi, N., and Garnett, R. (eds.), \emph{Proceedings of the 31st International Conference on Advances in Neural Information Processing Systems ({N}eur{IPS}'18)}, pp.\  6639–6649, 2018.

\bibitem[Qu et~al.(2025)Qu, Holzmüller, Varoquaux, and Morvan]{qu_tabicl_2025}
Qu, J., Holzmüller, D., Varoquaux, G., and Morvan, M.~L.
\newblock {TabICL}: {A} {Tabular} {Foundation} {Model} for {In}-{Context} {Learning} on {Large} {Data}, February 2025.
\newblock URL \url{http://arxiv.org/abs/2502.05564}.
\newblock arXiv:2502.05564 [cs].

\bibitem[Reuter et~al.(2025)Reuter, Rudner, Fortuin, and Rügamer]{reuter_can_2025}
Reuter, A., Rudner, T. G.~J., Fortuin, V., and Rügamer, D.
\newblock Can {Transformers} {Learn} {Full} {Bayesian} {Inference} in {Context}?, January 2025.
\newblock URL \url{http://arxiv.org/abs/2501.16825}.
\newblock arXiv:2501.16825 [cs].

\bibitem[Sanh et~al.(2020)Sanh, Debut, Chaumond, and Wolf]{sanh_distilbert_2020}
Sanh, V., Debut, L., Chaumond, J., and Wolf, T.
\newblock {DistilBERT}, a distilled version of {BERT}: smaller, faster, cheaper and lighter, March 2020.
\newblock URL \url{http://arxiv.org/abs/1910.01108}.
\newblock arXiv:1910.01108 [cs].

\bibitem[Shen et~al.(2020)Shen, Zhen, Ye, Ma, Yao, Gholami, Mahoney, and Keutzer]{shen_q-bert_2020}
Shen, S., Zhen, D., Ye, J., Ma, L., Yao, Z., Gholami, A., Mahoney, M., and Keutzer, K.
\newblock Q-{BERT}: {Hessian} {Based} {Ultra} {Low} {Precision} {Quantization} of {BERT}.
\newblock \emph{Proceedings of the AAAI Conference on Artificial Intelligence}, 34:\penalty0 8815--8821, April 2020.
\newblock \doi{10.1609/aaai.v34i05.6409}.

\bibitem[Teerapittayanon et~al.(2016)Teerapittayanon, McDanel, and Kung]{teerapittayanon_branchynet_2016}
Teerapittayanon, S., McDanel, B., and Kung, H.
\newblock {BranchyNet}: {Fast} inference via early exiting from deep neural networks.
\newblock In \emph{2016 23rd {International} {Conference} on {Pattern} {Recognition} ({ICPR})}, pp.\  2464--2469, December 2016.
\newblock \doi{10.1109/ICPR.2016.7900006}.
\newblock URL \url{https://ieeexplore.ieee.org/document/7900006}.

\bibitem[van Breugel \& van~der Schaar(2024)van Breugel and van~der Schaar]{vanbreugel-icml24a}
van Breugel, B. and van~der Schaar, M.
\newblock Why tabular foundation models should be a research priority.
\newblock In Salakhutdinov, R., Kolter, Z., Heller, K., Weller, A., Oliver, N., Scarlett, J., and Berkenkamp, F. (eds.), \emph{Proceedings of the 41st International Conference on Machine Learning ({ICML}'24)}, volume 251 of \emph{Proceedings of Machine Learning Research}. PMLR, 2024.

\bibitem[Vaswani et~al.(2017)Vaswani, Shazeer, Parmar, Uszkoreit, Jones, Gomez, Kaiser, and Polosukhin]{vaswani-neurips17a}
Vaswani, A., Shazeer, N., Parmar, N., Uszkoreit, J., Jones, L., Gomez, A., Kaiser, L., and Polosukhin, I.
\newblock Attention is all you need.
\newblock In  \citet{nips17}.

\bibitem[Wang et~al.(2020)Wang, Wei, Dong, Bao, Yang, and Zhou]{wang_minilm_2020}
Wang, W., Wei, F., Dong, L., Bao, H., Yang, N., and Zhou, M.
\newblock {MiniLM}: {Deep} {Self}-{Attention} {Distillation} for {Task}-{Agnostic} {Compression} of {Pre}-{Trained} {Transformers}.
\newblock In \emph{Advances in {Neural} {Information} {Processing} {Systems}}, volume~33, pp.\  5776--5788. Curran Associates, Inc., 2020.
\newblock URL \url{https://proceedings.neurips.cc/paper/2020/hash/3f5ee243547dee91fbd053c1c4a845aa-Abstract.html}.

\bibitem[Xie et~al.(2021)Xie, Raghunathan, Liang, and Ma]{xie_explanation_2021}
Xie, S.~M., Raghunathan, A., Liang, P., and Ma, T.
\newblock An {Explanation} of {In}-context {Learning} as {Implicit} {Bayesian} {Inference}.
\newblock October 2021.
\newblock URL \url{https://openreview.net/forum?id=RdJVFCHjUMI}.

\bibitem[Xin et~al.(2020)Xin, Tang, Lee, Yu, and Lin]{xin_deebert_2020}
Xin, J., Tang, R., Lee, J., Yu, Y., and Lin, J.
\newblock {DeeBERT}: {Dynamic} {Early} {Exiting} for {Accelerating} {BERT} {Inference}.
\newblock In Jurafsky, D., Chai, J., Schluter, N., and Tetreault, J. (eds.), \emph{Proceedings of the 58th {Annual} {Meeting} of the {Association} for {Computational} {Linguistics}}, pp.\  2246--2251, Online, July 2020. Association for Computational Linguistics.
\newblock \doi{10.18653/v1/2020.acl-main.204}.
\newblock URL \url{https://aclanthology.org/2020.acl-main.204/}.

\bibitem[Zafrir et~al.(2019)Zafrir, Boudoukh, Izsak, and Wasserblat]{zafrir_q8bert_2019}
Zafrir, O., Boudoukh, G., Izsak, P., and Wasserblat, M.
\newblock {Q8BERT}: {Quantized} {8Bit} {BERT}.
\newblock In \emph{2019 {Fifth} {Workshop} on {Energy} {Efficient} {Machine} {Learning} and {Cognitive} {Computing} - {NeurIPS} {Edition} ({EMC2}-{NIPS})}, pp.\  36--39, December 2019.
\newblock \doi{10.1109/EMC2-NIPS53020.2019.00016}.
\newblock URL \url{https://ieeexplore.ieee.org/abstract/document/9463531}.

\end{thebibliography}
\bibliographystyle{icml}

\newpage
\appendix
\onecolumn

\section{Decoder Pre-Training.}\label{app:decoder_training}
We pre-train each decoder independently on synthetic datasets sampled from the prior, following the prior setup from \citep{hollmann-iclr23a}. During training, all Transformer components are frozen; only the target decoder is updated. Each decoder is attached at a specific Transformer layer, receiving its output as input, and trained for a total of 100 epochs. Detailed training settings are provided in Table ~\ref{tab:training_params}. Settings for the synthetic datasets from the prior are stated in Table~\ref{tab:prior_params}. We train all decoders on a single NVIDIA RTX 2080 GPU.
\begin{table*}[h]
    \begin{minipage}{.45\linewidth}
    \centering
    \caption{\textbf{Training Parameters.} We report the parameters for the pre-training setup. We train each decoder for a total of 100 epochs. \#Steps/Epoch states the number of synthetically drawn datasets per epoch from the prior.}
    \label{tab:training_params}
    \begin{tabular}{l|c}
    \toprule
    \textbf{Training Parameters} & Value \\
    \midrule
         \textbf{Epochs} & 100 \\
         \textbf{Batch Size} & 8 \\
         \textbf{\#Steps/Epoch} & 1024\\
         \textbf{Learning Rate} & 3e-5 \\
    \bottomrule
    \end{tabular}
    \end{minipage}
    \hspace{1cm}
    \begin{minipage}{.45\linewidth}
    \centering
    \caption{\textbf{Parameters for the Prior.} We report the parameters for the synthetic datasets drawn from the Prior.}
    \label{tab:prior_params}
    \begin{tabular}{l|c}
    \toprule
    \textbf{Prior Parameters} & Value \\
    \midrule
        \textbf{\#Samples per Dataset} & 1152 \\
        \textbf{\#Features per Dataset} &  100 \\
        \textbf{\#Max Classes per Dataset} & 10 \\
    \bottomrule
    \end{tabular}
    \end{minipage}
\end{table*}

\clearpage
\newpage
\section{Algorithm Details.}\label{app:alg_details}
Algorithm~\ref{alg:earlyexit} specifies the implementation of dynamic early exiting in tabular in-context learning.
\begin{algorithm}[]
   \caption{Early-Exit Inference for TabPFN}
   \label{alg:earlyexit}
\begin{algorithmic}
   \STATE {\bfseries Input:} Table embedding $x = (x_{\text{train}}, y_{\text{train}}, x_{\text{test}})$, threshold $\tau$
   \FOR{$i = 1$ to $N_{Transformer Layers}$}
   \STATE $x \leftarrow \text{TransformerLayer}_i(x)$
   \STATE $\hat{y} \leftarrow \text{Decoder}_i(x_{\text{test}})$
   \STATE $p \leftarrow \text{softmax}(\hat{y})$
   \STATE $H \leftarrow -\sum p \log p$
   \IF{$H < \tau$}
       \STATE \textbf{return} $p$
   \ENDIF
   \ENDFOR
   \STATE \textbf{return} $p$
\end{algorithmic}
\end{algorithm}

\clearpage
\newpage
\section{Evaluation Details.}\label{app:eval_details}
\subsection{Datasets}\label{app:datasets}
We report details about the used small datasets in Table~\ref{tab:pmlb_datasets} and large datasets in Table~\ref{tab:nature_datasets}.
\begin{table}[h]
    \centering
    \caption{\textbf{Small Datasets.} The datasets used from the pmlb-mini suite~\citep{knauer_pmlbmini_2024}}
    \begin{tabular}{l|ccccc}
    \toprule
    Dataset & OpenML ID & \#Features & \#Samples & \#Targets \\
    \midrule
    parity5 & 40714 & 5 & 32 & 2 \\
    analcatdata\_fraud & 40660 & 11 & 42 & 2 \\
    aids & 346 & 4 & 50 & 2 \\
    analcatdata\_bankruptcy & 476 & 6 & 50 & 2 \\
    analcatdata\_japansolvent & 467 & 9 & 52 & 2 \\
    analcatdata\_asbestos & 459 & 3 & 83 & 2 \\
    lupus & 472 & 3 & 87 & 2 \\
    postoperative-patient-data & 40683 & 8 & 88 & 2\\
    analcatdata\_cyyoung9302 & 479 & 10 & 92 & 2\\
    analcatdata\_cyyoung8092 & 465 & 10 & 97 & 2 \\
    analcatdata\_creditscore & 461 & 6 & 100 & 2 \\
    molecular-biology\_promoters & 956 & 57 & 106 & 2\\
    analcatdata\_boxing1 & 448 & 3 & 120 & 2 \\
    mux6 & 40681 & 6 & 128 & 2 \\
    analcatdata\_boxing2 & 444 & 3 & 132 & 2 \\
    corral & 40669 & 6 & 160 & 2 \\
    backache & 463 & 32 & 180 & 2 \\
    prnn\_crabs & 446 & 7 & 200 & 2 \\
    sonar & 40 & 60 & 208 & 2 \\
    biomed & 481 & 8 & 209 & 2 \\
    prnn\_synth & 464 & 2 & 250 & 2\\
    analcatdata\_lawsuit & 450 & 4 & 264 & 2 \\
    SPECT & 336 & 22 & 267 & 2 \\
    heart-statlog & 53 & 13 & 270 & 2\\
    hepatitis & 55 & 19 & 155 & 2 \\
    breast-cancer & 13 & 9 & 286 & 2 \\
    hungarian & 231 & 13 & 294 & 2 \\
    cleve & 40710 & 13 & 303 & 2 \\
    haberman & 43 & 3 & 306 & 2 \\
    SPECTF & 337 & 44 & 349 & 2 \\
    ionosphere & 59 & 34 & 351 & 2 \\
    colic & 27 & 22 & 368 & 2 \\
    vote & 56 & 16 & 435 & 2 \\
    irish & 451 & 5 & 500 & 2 \\
    \bottomrule
    \end{tabular}
    \label{tab:pmlb_datasets}
\end{table}

\begin{table}[h]
    \centering
    \caption{\textbf{Large Datasets.} The datasets used from the original TabPFNv2 evaluation.~\citep{hollmann-nature25a}}
    \begin{tabular}{l|ccccc}
    \toprule
    Dataset & OpenML ID & \#Features & \#Samples & \#Targets \\
    \midrule
    ada & 41156 & 48 & 4147 & 2 \\
    churn & 49791 & 20 & 5000 & 2 \\
    phoneme & 1489 & 5 & 5404 & 2 \\
    Satellite & 40900 & 36 & 5100 & 2 \\
    sylvine & 41146 & 20 & 5124 & 2 \\
    \bottomrule
    \end{tabular}
    \label{tab:nature_datasets}
\end{table}

\subsection{Hyperparameters}\label{app:hyperparameters}
We use TabPFNv2 \citep{hollmann-nature25a} for all experiments. We use default classification hyperparameters from the paper evaluation.

\clearpage
\newpage
\section{Additional Results for Large Datasets and Small Datasets.}
\begin{figure}[h]
    \centering
    \includegraphics[width=1.0\linewidth]{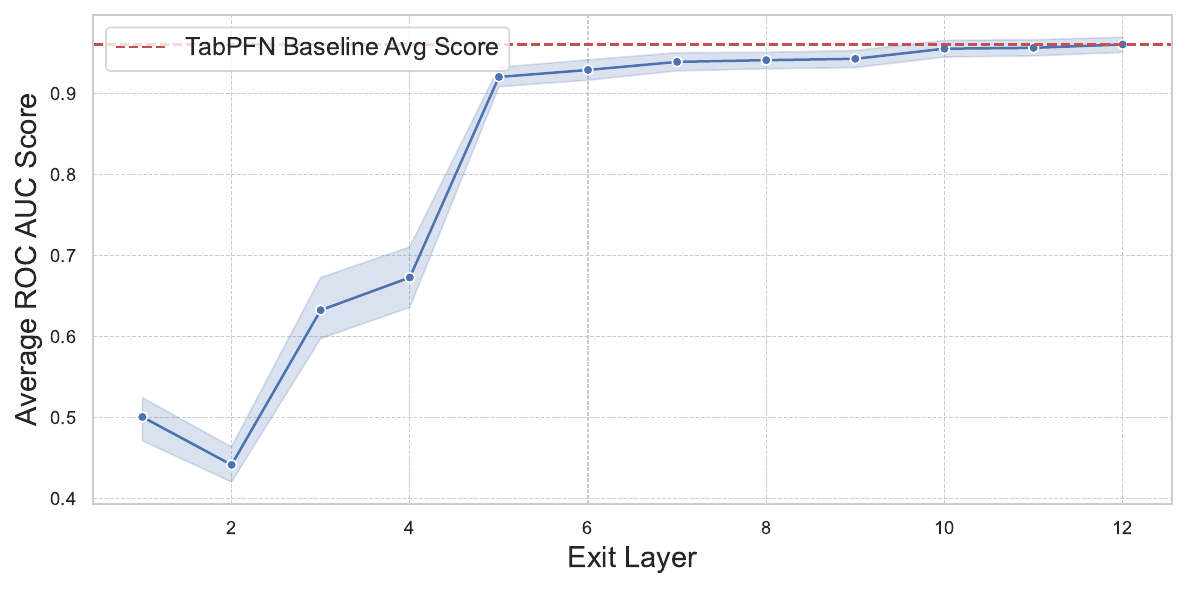}
    \caption{\textbf{Performance per Layer over 5 Large Datasets with Individual Decoders.} We present the performance average ROC AUC score over 5 large datasets with up to 5k samples averaged over 10 folds as well as the 95\% confidence interval. Additionally we present the baseline of TabPFN when passing through the entire transformer in red. Similar to small datasets, early layers already exhibit strong performance.}
    \label{fig:nature-exit-layers}
\end{figure}

We present evaluation results—similar to the evaluation of the small datasets—for large datasets. Figure~\ref{fig:pmlb-dynamic-stoping} illustrates the tradeoff between relative runtime improvement and performance decrease for small datasets. Figure~\ref{fig:nature-exit-layers} shows the performance per layer over large datasets.
\begin{figure}[h]
    \centering
    \includegraphics[width=1.0\linewidth]{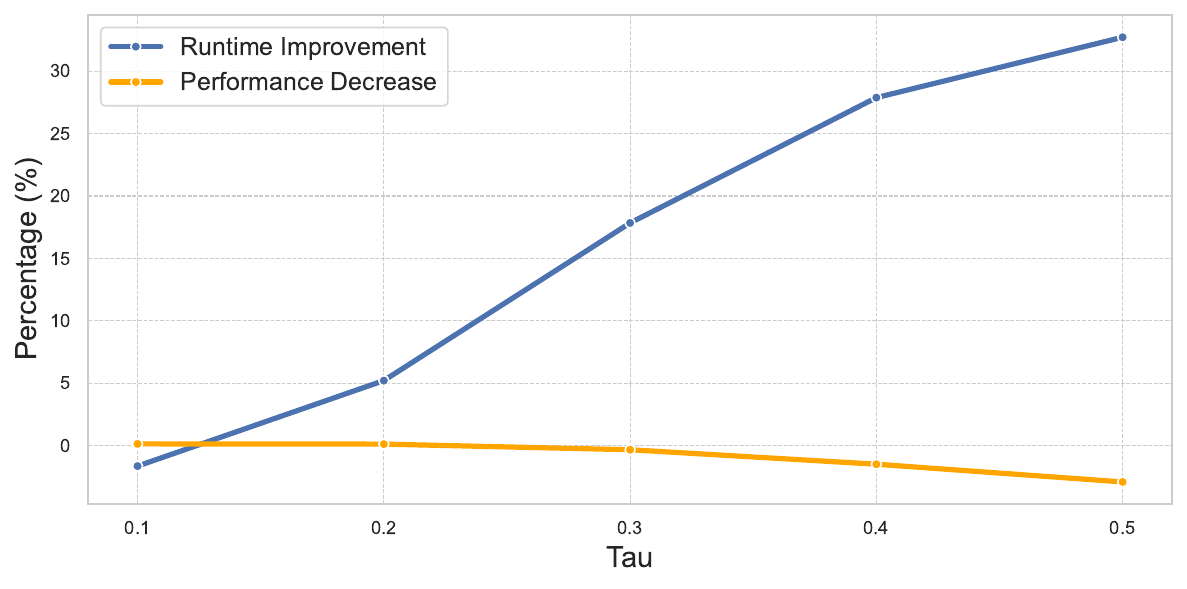}
    \caption{\textbf{Tradeoff between Improved Runtime and Decrease in Predictive Performance.} We report the relative improvements in runtime as well as the relative decrease in ROC AUC score averaged over 34 small datasets when early stopping based on entropy for 5 different entropy thresholds.}
    \label{fig:pmlb-dynamic-stopping}
\end{figure}

\clearpage
\newpage
\section{Detailed Results.}\label{app:detailed_results}
We present the per-dataset results of our evaluation of small datasets in Table~\ref{tab:full_scores_pmlb} and for large datasets in Table~\ref{tab:full_scores_nature}.
We further report the exit layer per dataset, for small datasets in Table~\ref{tab:exit_layers_pmlb} and for large datasets in Table~\ref{tab:exit_layers_nature}.
\begin{table}[]
    \centering
    \caption{\textbf{ROC AUC Scores per Dataset over Different Thresholds for Small Datasets.} We report ROC AUC scores per dataset for 5 different thresholds. All scores are averages over 10 fold cross validation.}
    \begin{tabular}{l|ccccc}
    \toprule
    Dataset & $\tau=0.1$ & $\tau=0.2$ & $\tau=0.3$ & $\tau=0.4$ & $\tau=0.5$ \\
    \midrule
    parity5 & $1.000$  $\pm$ $0.00$ &$1.000$  $\pm$ $0.00$ &$1.000$  $\pm$ $0.00$ &$0.950$  $\pm$ $0.16$ &$0.725$  $\pm$ $0.38$ \\
    analcatdata\_fraud & $0.833$  $\pm$ $0.24$ &$0.867$  $\pm$ $0.23$ &$0.800$  $\pm$ $0.28$ &$0.733$  $\pm$ $0.31$ &$0.767$  $\pm$ $0.27$ \\
    aids & $1.000$  $\pm$ $0.00$ &$1.000$  $\pm$ $0.00$ &$1.000$  $\pm$ $0.00$ &$0.850$  $\pm$ $0.32$ &$0.783$  $\pm$ $0.39$ \\
    analcatdata\_bankruptcy & $0.967$  $\pm$ $0.11$ &$0.967$  $\pm$ $0.11$ &$0.967$  $\pm$ $0.11$ &$0.967$  $\pm$ $0.11$ &$0.950$  $\pm$ $0.16$ \\
    analcatdata\_japansolvent & $0.933$  $\pm$ $0.14$ &$0.950$  $\pm$ $0.11$ &$0.900$  $\pm$ $0.22$ &$0.900$  $\pm$ $0.22$ &$0.900$  $\pm$ $0.22$ \\
    analcatdata\_asbestos & $0.871$  $\pm$ $0.15$ &$0.871$  $\pm$ $0.15$ &$0.872$  $\pm$ $0.15$ &$0.866$  $\pm$ $0.15$ &$0.866$  $\pm$ $0.15$ \\
    lupus &  $0.820$  $\pm$ $0.18$ &$0.820$  $\pm$ $0.18$ &$0.826$  $\pm$ $0.17$ &$0.809$  $\pm$ $0.18$ &$0.728$  $\pm$ $0.24$ \\
    postoperative-patient-data & $0.356$  $\pm$ $0.14$ &$0.356$  $\pm$ $0.14$ &$0.356$  $\pm$ $0.14$ &$0.384$  $\pm$ $0.15$ &$0.448$  $\pm$ $0.19$ \\
    analcatdata\_cyyoung9302 & $0.917$  $\pm$ $0.16$ &$0.903$  $\pm$ $0.16$ &$0.896$  $\pm$ $0.16$ &$0.915$  $\pm$ $0.16$ &$0.915$  $\pm$ $0.16$ \\
    analcatdata\_cyyoung8092 & $0.880$  $\pm$ $0.13$ &$0.896$  $\pm$ $0.12$ &$0.883$  $\pm$ $0.14$ &$0.864$  $\pm$ $0.14$ &$0.864$  $\pm$ $0.14$ \\
    analcatdata\_creditscore & $1.000$  $\pm$ $0.00$ &$1.000$  $\pm$ $0.00$ &$1.000$  $\pm$ $0.00$ &$0.995$  $\pm$ $0.02$ &$0.995$  $\pm$ $0.02$ \\
    molecular-biology\_promoters & $0.579$  $\pm$ $0.25$ &$0.585$  $\pm$ $0.25$ &$0.589$  $\pm$ $0.27$ &$0.610$  $\pm$ $0.24$ &$0.601$  $\pm$ $0.24$ \\
    analcatdata\_boxing1 & $0.879$  $\pm$ $0.10$ &$0.879$  $\pm$ $0.10$ &$0.882$  $\pm$ $0.10$ &$0.892$  $\pm$ $0.10$ &$0.882$  $\pm$ $0.10$ \\
    mux6 &  $1.000$  $\pm$ $0.00$ &$1.000$  $\pm$ $0.00$ &$0.990$  $\pm$ $0.03$ &$0.943$  $\pm$ $0.10$ &$0.904$  $\pm$ $0.11$ \\
    analcatdata\_boxing2 & $0.853$  $\pm$ $0.09$ &$0.855$  $\pm$ $0.09$ &$0.853$  $\pm$ $0.09$ &$0.863$  $\pm$ $0.08$ &$0.822$  $\pm$ $0.10$ \\
    corral &  $1.000$  $\pm$ $0.00$ &$0.987$  $\pm$ $0.02$ &$0.979$  $\pm$ $0.04$ &$0.976$  $\pm$ $0.04$ &$0.976$  $\pm$ $0.04$ \\
    backache &  $0.701$  $\pm$ $0.13$ &$0.711$  $\pm$ $0.15$ &$0.703$  $\pm$ $0.11$ &$0.700$  $\pm$ $0.11$ &$0.697$  $\pm$ $0.11$ \\
    prnn\_crabs &  $1.000$  $\pm$ $0.00$ &$1.000$  $\pm$ $0.00$ &$0.999$  $\pm$ $0.00$ &$0.999$  $\pm$ $0.00$ &$0.999$  $\pm$ $0.00$ \\
    sonar &  $0.944$  $\pm$ $0.05$ &$0.944$  $\pm$ $0.05$ &$0.944$  $\pm$ $0.05$ &$0.920$  $\pm$ $0.09$ &$0.902$  $\pm$ $0.09$ \\
    biomed &  $1.000$  $\pm$ $0.00$ &$0.978$  $\pm$ $0.02$ &$0.971$  $\pm$ $0.03$ &$0.973$  $\pm$ $0.03$ &$0.973$  $\pm$ $0.03$ \\
    prnn\_synth &  $0.948$  $\pm$ $0.03$ &$0.945$  $\pm$ $0.03$ &$0.946$  $\pm$ $0.03$ &$0.946$  $\pm$ $0.03$ &$0.946$  $\pm$ $0.03$ \\
    analcatdata\_lawsuit &  $0.990$  $\pm$ $0.02$ &$0.990$  $\pm$ $0.02$ &$0.990$  $\pm$ $0.02$ &$0.990$  $\pm$ $0.02$ &$0.990$  $\pm$ $0.02$ \\
    SPECT &  $0.833$  $\pm$ $0.07$ &$0.834$  $\pm$ $0.07$ &$0.844$  $\pm$ $0.08$ &$0.848$  $\pm$ $0.08$ &$0.848$  $\pm$ $0.08$ \\
    heart-statlog &  $0.908$  $\pm$ $0.06$ &$0.908$  $\pm$ $0.06$ &$0.912$  $\pm$ $0.06$ &$0.911$  $\pm$ $0.06$ &$0.910$  $\pm$ $0.06$ \\
    hepatitis &  $0.856$  $\pm$ $0.09$ &$0.864$  $\pm$ $0.09$ &$0.879$  $\pm$ $0.08$ &$0.876$  $\pm$ $0.10$ &$0.879$  $\pm$ $0.10$ \\
    breast-cancer &  $0.728$  $\pm$ $0.06$ &$0.730$  $\pm$ $0.06$ &$0.729$  $\pm$ $0.06$ &$0.738$  $\pm$ $0.07$ &$0.730$  $\pm$ $0.07$ \\
    hungarian &  $0.908$  $\pm$ $0.06$ &$0.909$  $\pm$ $0.06$ &$0.909$  $\pm$ $0.07$ &$0.912$  $\pm$ $0.07$ &$0.912$  $\pm$ $0.07$ \\
    cleve &  $0.910$  $\pm$ $0.05$ &$0.911$  $\pm$ $0.05$ &$0.912$  $\pm$ $0.06$ &$0.909$  $\pm$ $0.06$ &$0.909$  $\pm$ $0.06$ \\
    haberman &  $0.714$  $\pm$ $0.17$ &$0.714$  $\pm$ $0.17$ &$0.719$  $\pm$ $0.16$ &$0.700$  $\pm$ $0.13$ &$0.697$  $\pm$ $0.12$ \\
    SPECTF &  $0.964$  $\pm$ $0.04$ &$0.964$  $\pm$ $0.04$ &$0.951$  $\pm$ $0.05$ &$0.949$  $\pm$ $0.04$ &$0.949$  $\pm$ $0.04$ \\
    ionosphere &  $0.983$  $\pm$ $0.02$ &$0.974$  $\pm$ $0.02$ &$0.957$  $\pm$ $0.04$ &$0.949$  $\pm$ $0.04$ &$0.949$  $\pm$ $0.04$ \\
    colic &  $0.901$  $\pm$ $0.04$ &$0.900$  $\pm$ $0.04$ &$0.892$  $\pm$ $0.04$ &$0.868$  $\pm$ $0.05$ &$0.869$  $\pm$ $0.05$ \\
    vote &  $0.992$  $\pm$ $0.01$ &$0.983$  $\pm$ $0.02$ &$0.979$  $\pm$ $0.03$ &$0.979$  $\pm$ $0.03$ &$0.979$  $\pm$ $0.03$ \\
    irish &  $1.000$  $\pm$ $0.00$ &$1.000$  $\pm$ $0.00$ &$1.000$  $\pm$ $0.00$ &$1.000$  $\pm$ $0.00$ &$1.000$  $\pm$ $0.00$  \\
    \bottomrule
    \end{tabular}
    \label{tab:full_scores_pmlb}
\end{table}

\begin{table}[]
    \centering
    \caption{\textbf{Exit Layer per Dataset over Different Thresholds for Small Datasets.} We report the exit layer per dataset for 5 different thresholds when dynamically early stopping. All exit layers are averages over 10 fold cross validation.}
    \begin{tabular}{l|ccccc}
    \toprule
    Dataset & $\tau=0.1$ & $\tau=0.2$ & $\tau=0.3$ & $\tau=0.4$ & $\tau=0.5$ \\
    \midrule
    parity5 & $10.3$ & $10.0$ & $9.4$ & $7.1$ & $5.3$ \\
    analcatdata\_fraud & $12.0$ & $10.6$ & $7.1$ & $5.7$ & $5.0$ \\
    aids & $12.0$ & $12.0$ & $12.0$ & $11.2$ & $9.1$ \\
    analcatdata\_bankruptcy & $11.9$ & $9.8$ & $6.7$ & $5.7$ & $5.0$ \\
    analcatdata\_japansolvent & $12.0$ & $12.0$ & $6.7$ & $5.6$ & $5.0$ \\
    analcatdata\_asbestos & $12.0$ & $11.6$ & $7.2$ & $5.0$ & $5.0$ \\
    lupus & $12.0$ & $12.0$ & $11.3$ & $7.9$ & $5.0$ \\
    postoperative-patient-data & $12.0$ & $12.0$ & $12.0$ & $10.6$ & $5.7$ \\
    analcatdata\_cyyoung9302 & $11.3$ & $7.5$ & $5.2$ & $5.0$ & $5.0$ \\
    analcatdata\_cyyoung8092 & $12.0$ & $10.8$ & $5.9$ & $5.0$ & $5.0$ \\
    analcatdata\_creditscore & $8.0$ & $6.0$ & $5.0$ & $5.0$ & $5.0$ \\
    molecular-biology\_promoters & $12.0$ & $12.0$ & $11.3$ & $7.1$ & $5.0$ \\
    analcatdata\_boxing1 & $12.0$ & $12.0$ & $11.6$ & $7.7$ & $5.0$ \\
    mux6 & $11.9$ & $11.4$ & $10.6$ & $8.5$ & $5.9$ \\
    analcatdata\_boxing2 & $12.0$ & $12.0$ & $12.0$ & $11.0$ & $7.8$ \\
    corral & $9.5$ & $5.9$ & $5.0$ & $5.0$ & $5.0$ \\
    backache & $12.0$ & $8.1$ & $5.0$ & $5.0$ & $5.0$ \\
    prnn\_crabs & $7.5$ & $6.1$ & $5.0$ & $5.0$ & $5.0$ \\
    sonar & $12.0$ & $12.0$ & $10.9$ & $7.0$ & $5.2$ \\
    biomed & $9.6$ & $6.5$ & $5.1$ & $5.0$ & $5.0$ \\
    prnn\_synth & $12.0$ & $10.8$ & $5.0$ & $5.0$ & $5.0$ \\
    analcatdata\_lawsuit & $6.1$ & $5.0$ & $5.0$ & $5.0$ & $5.0$ \\
    SPECT & $12.0$ & $10.3$ & $5.7$ & $5.0$ & $5.0$ \\
    heart-statlog & $12.0$ & $11.4$ & $7.0$ & $5.0$ & $5.0$ \\
    hepatitis & $12.0$ & $6.8$ & $5.0$ & $5.0$ & $5.0$ \\
    breast-cancer & $12.0$ & $12.0$ & $12.0$ & $8.2$ & $5.1$ \\
    hungarian & $12.0$ & $10.0$ & $5.1$ & $5.0$ & $5.0$ \\
    cleve & $12.0$ & $12.0$ & $6.1$ & $5.0$ & $5.0$ \\
    haberman & $12.0$ & $12.0$ & $10.7$ & $5.4$ & $5.0$ \\
    SPECTF & $12.0$ & $12.0$ & $9.2$ & $5.0$ & $5.0$ \\
    ionosphere & $11.3$ & $6.7$ & $5.3$ & $5.0$ & $5.0$ \\
    colic & $12.0$ & $12.0$ & $8.4$ & $5.0$ & $5.0$ \\
    vote & $6.8$ & $5.1$ & $5.0$ & $5.0$ & $5.0$ \\
    irish & $6.4$ & $5.3$ & $5.0$ & $5.0$ & $5.0$ \\
    \bottomrule
    \end{tabular}
    \label{tab:exit_layers_pmlb}
\end{table}

\begin{table}[]
    \centering
    \caption{\textbf{ROC AUC Scores per Dataset over Different Thresholds for Small Datasets.} We report ROC AUC scores per dataset for 5 different thresholds. All scores are averages over 10 fold cross validation.}
    \begin{tabular}{l|ccccc}
    \toprule
    Dataset & $\tau=0.1$ & $\tau=0.2$ & $\tau=0.3$ & $\tau=0.4$ & $\tau=0.5$ \\
    \midrule
    ada &$0.919$ $\pm$ $0.02$ &$0.908$ $\pm$ $0.02$ &$0.9$ $\pm$ $0.02$ &$0.899$ $\pm$ $0.02$ &$0.899$ $\pm$ $0.02$ \\
    churn &$0.919$ $\pm$ $0.03$ &$0.885$ $\pm$ $0.03$ &$0.886$ $\pm$ $0.03$ &$0.888$ $\pm$ $0.03$ &$0.888$ $\pm$ $0.03$ \\
    phoneme &$0.969$ $\pm$ $0.01$ &$0.969$ $\pm$ $0.01$ &$0.89$ $\pm$ $0.02$ &$0.877$ $\pm$ $0.02$ &$0.877$ $\pm$ $0.02$ \\
    Satellite &$0.987$ $\pm$ $0.02$ &$0.972$ $\pm$ $0.02$ &$0.972$ $\pm$ $0.02$ &$0.972$ $\pm$ $0.02$ &$0.972$ $\pm$ $0.02$ \\
    sylvine &$0.992$ $\pm$ $0.0$ &$0.971$ $\pm$ $0.0$ &$0.968$ $\pm$ $0.01$ &$0.965$ $\pm$ $0.01$ &$0.965$ $\pm$ $0.01$ \\
    \bottomrule
    \end{tabular}
    \label{tab:full_scores_nature}
\end{table}

\begin{table}[]
    \centering
    \caption{\textbf{Exit Layer per Dataset over Different Thresholds for Large Datasets.} We report the exit layer per dataset for 5 different thresholds when dynamically early stopping. All exit layers are averages over 10 fold cross validation.}
    \begin{tabular}{l|ccccc}
    \toprule
    Dataset & $\tau=0.1$ & $\tau=0.2$ & $\tau=0.3$ & $\tau=0.4$ & $\tau=0.5$ \\
    \midrule
    ada & $12.0$ & $6.0$ & $5.1$ & $5.0$ & $5.0$ \\
    churn & $9.5$ & $6.0$ & $5.5$ & $5.0$ & $5.0$ \\
    phoneme & $12.0$ & $12.0$ & $6.0$ & $5.0$ & $5.0$ \\
    Satellite & $5.9$ & $5.0$ & $5.0$ & $5.0$ & $5.0$ \\
    sylvine & $8.0$ & $6.1$ & $5.5$ & $5.0$ & $5.0$ \\
    \bottomrule
    \end{tabular}
    \label{tab:exit_layers_nature}
\end{table}

\end{document}